\documentclass[sigplan,10pt]{acmart}

\renewcommand\footnotetextcopyrightpermission[1]{}

\usepackage{tikz}
\usepackage{amsmath}
\usepackage{enumitem}
\usepackage{array}
\usepackage{booktabs}
\usepackage{pifont}
\usepackage{subcaption}
 \usepackage{svg}
\usepackage{enumitem}
\usepackage[ruled,linesnumbered]{algorithm2e}
\hypersetup{colorlinks=true, allcolors=black}
\author{Muyang Cao}
\affiliation{
  \institution{Zhejiang University}
  \city{Hangzhou}
  \country{China}
}
\email{muyangc@zju.edu.cn}

\author{Di Yu}
\affiliation{
  \institution{Zhejiang University}
  \city{Hangzhou}
  \country{China}
}
\email{yudi2023@zju.edu.cn}

\author{Peiling Zhu}
\affiliation{
  \institution{Zhejiang University}
  \city{Hangzhou}
  \country{China}
}
\email{3250100550@zju.edu.cn}

\author{Linshan Jiang}
\affiliation{
  \institution{Southern University of Science and Technology}
  \city{Shenzhen}
  \country{China}
}
\email{jiangls@mail.sustech.edu.cn}

\author{Xin Du}
\authornote{Corresponding authors.}
\affiliation{
  \institution{Zhejiang University}
  \city{Hangzhou}
  \country{China}
}
\email{xindu@zju.edu.cn}

\author{Shuiguang Deng}
\affiliation{
  \institution{Zhejiang University}
  \city{Hangzhou}
  \country{China}
}
\email{dengsg@zju.edu.cn}

\author{Gang Pan}
\authornotemark[1]
\affiliation{
  \institution{Zhejiang University}
  \city{Hangzhou}
  \country{China}
}
\email{gpan@zju.edu.cn}

\begin{document}
\pagestyle{plain}

\date{}

\title{Migrate Computation, Not Data: Fault Recovery in Neuromorphic Many-Core Systems}


\setlist{topsep=2pt}     
\setlength{\textfloatsep}{8pt}       
\setlength{\abovecaptionskip}{4pt}   
\setlength{\belowcaptionskip}{2pt}   
\setlength{\abovedisplayskip}{6pt}   
\setlength{\belowdisplayskip}{6pt}   
\makeatletter
\makeatother

\begin{abstract}

As neuromorphic systems scale to thousands of near-memory cores, aging-induced permanent faults in neuron-update datapaths become an important obstacle to long-term deployment. Existing online recovery methods typically retire a faulty core and migrate its neurons, along with their weights and membrane states, to healthy cores. This whole-core treatment incurs high recovery traffic and makes recovery depend on scarce destinations; at high utilization, there may be enough surviving compute capacity but insufficient free storage to copy the affected neuron context. This paper presents SpikeRelay, an online recovery framework for diagnosed compute-unit faults. SpikeRelay exploits the fact that, in many neuromorphic cores, the compute datapath, local SRAM, and router are distinct modules: a compute fault need not destroy stored weights and states or disconnect the core from the NoC. Instead of retiring a faulty core and migrating its neuron data, SpikeRelay converts the faulty core into a storage-and-forwarding node while offloading neuron-update computation to a healthy rescue core. This decoupling of storage and computation eliminates recovery-time data migration and greatly reduces spare-SRAM requirements. Consequently, SpikeRelay sustains 100\% recovery at 92.2\% SRAM utilization and a 50\% fault rate, whereas the best baseline achieves only 22.1\%.

\end{abstract}

\maketitle



\section{Introduction}
Neuromorphic processors execute spiking neural networks (SNNs) on arrays of near-memory cores, where each core integrates a compute unit, local SRAM, and a Network-on-Chip (NoC) router~\cite{roy2019towards,davies2018loihi}. Such architectures are well suited to always-on edge, robotic, and space-constrained workloads~\cite{ma2022neuromorphic,ortiz2024energy,bartolozzi2022embodied}, as neuron states and synaptic data remain local to computation, while event-driven processing exploits sparse one-bit spike communication to achieve high energy efficiency~\cite{yu2026masi}. As these systems scale from hundreds to thousands of cores and are deployed for long service lifetimes, however, permanent hardware faults become an unavoidable architectural concern~\cite{putra2023rescuesnn,navaridas2013spinnaker}. Taking such systems offline for repair is often impractical, so recovery must preserve execution while the system remains in service.

Against this backdrop, this paper focuses on a specific but important fault class: diagnosed, in-field permanent faults~\cite{putra2023rescuesnn,furber2014spinnaker,navaridas2013spinnaker} in the compute unit that performs neuron-state updates. Such faults can corrupt the execution datapath of the spiking neuron model, including membrane potential integration, leakage and reset~\cite{borkar2006designing,srinivasan2004impact,song2020aging,titirsha2021endurance}. Unlike transient perturbations, which affect an operation and then disappear, a permanent compute unit fault repeatedly applies an incorrect state-transition function whenever the affected neurons are updated. This distinction changes the recovery target: the system must restore the affected neurons' state-transition behavior, rather than merely mask an isolated erroneous output~\cite{putra2023rescuesnn,furber2014spinnaker}.

\begin{figure}[t]
  \centering
  \includegraphics[width=\columnwidth]{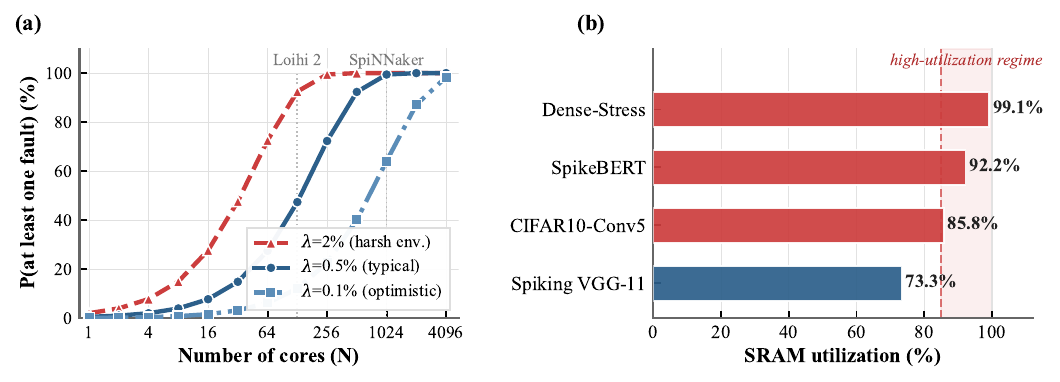}
  \caption{Faults are inevitable exactly where recovery has no room to move data. (a)~As core count grows, the probability of at least one in-field permanent fault approaches certainty. (b)~Most evaluated SNN workloads already run in the high-utilization regime ($\geq$85\%), leaving little spare SRAM for post-fault relocation.}
  \label{fig:motivation}
\end{figure}

At the scale of modern many-core neuromorphic processors, even a tiny per-core fault rate compounds across thousands of cores into a near-certainty\cite{srinivasan2004impact}. Consider TrueNorth's 4{,}096-core system~\cite{merolla2014million}, where a mere 0.1\% per-core rate already makes at least one permanent compute-unit fault 98\% likely, and a 0.5\% rate renders it all but inevitable (Figure~\ref{fig:motivation}(a)). Once such a fault occurs, SNN dynamics can amplify its effect because the membrane potential carries temporal context across time steps. Consistent with this sensitivity, prior work shows that only 0.012\% faulty processing elements can halve accuracy on a systolic accelerator~\cite{siddique2023improving}; in our multi-core neuromorphic experiments, test accuracy on the CIFAR-10 dataset drops by more than half at a 30\% fault rate and approaches random guessing beyond 40\%. Yet lightweight online recovery for permanent compute-unit faults remains largely unexplored in neuromorphic architectures.

Part of the reason is that a viable recovery scheme has to satisfy three coupled demands at once. First, it must restore the computational function of the faulty core so that its assigned neurons continue to execute correctly. Second, it must perform the repair while the chip keeps running in the field, because the membrane state cannot be rebuilt once it is lost and the chip is rarely allowed to go offline for maintenance. Third, it must remain effective when spare on-chip capacity is scarce, as mapping tools pack neurons tightly and real deployments run at high SRAM utilization~\cite{balaji2019mapping}. As Figure~\ref{fig:motivation}(b) shows, our evaluated workloads already operate deep inside this high-utilization regime, leaving little free SRAM into which a faulty core's neurons could be relocated\cite{fang2023spikingjelly,sengupta2019going,wu2019direct}.

Prior recovery techniques do not simultaneously satisfy the requirements of online execution, high utilization, and permanent compute-fault recovery. Software compensation, such as retraining or threshold adjustment, can improve fault tolerance for some models, but it is typically offline or model-specific and does not restore the failed compute function~\cite{siddique2023improving}. Hardware redundancy provides fast recovery when sufficient spare compute and memory resources are available~\cite{lyons1962use,yousuf2025layer}. Such schemes are effective below their provisioned budget, but once accumulated faults exceed that budget, or when spare cores lack sufficient SRAM to host the migrated neuron context. Runtime migration is more flexible, but by copying weights and states it turns recovery into a destination-SRAM allocation problem and leaves the faulty core’s surviving SRAM stranded.

Our key observation is that whole-core retirement is stronger than necessary for diagnosed compute-unit faults. In many neuromorphic cores, the Compute Unit, local SRAM, and NoC router are architecturally distinct\cite{davies2018loihi,ma2024darwin3}. A fault localized to the compute unit does not by itself destroy the stored weights and membrane states, nor does it disconnect the core from the network. If post-fault diagnosis confirms that storage and routing remain usable, the faulty core can continue to serve as the storage and communication endpoint for its neurons. The missing function is only the state-transition computation. Recovery can therefore be reframed as a task-rebinding problem: execute the LIF update on another core while preserving the original state, input, write-back, and spike-output paths.

%
Building on this observation, we design SpikeRelay, which migrates computation rather than model data.In SpikeRlay, after a compute unit fault, the faulty core is degraded to a storage-and-forwarding node, while a healthy rescue core executes the affected neuron updates. The rescue core need not be an idle spare and does not store the weights or membrane states of the faulty core. Instead, it discovers their original addresses from a fixed SRAM metadata layout, receives the same incoming spike events through a forwarding rule at the faulty router, accumulates the same per-timestep weighted inputs, remotely reads the required state, performs the LIF update locally, writes the updated state back to the faulty core, and emits output spikes with the original logical neuron identity. Thus, affected neurons keep their data and identity anchored at the faulty core, while only their computation is re-routed to a healthy datapath.

\begin{figure}[t]
  \centering
  \includegraphics[width=\columnwidth]{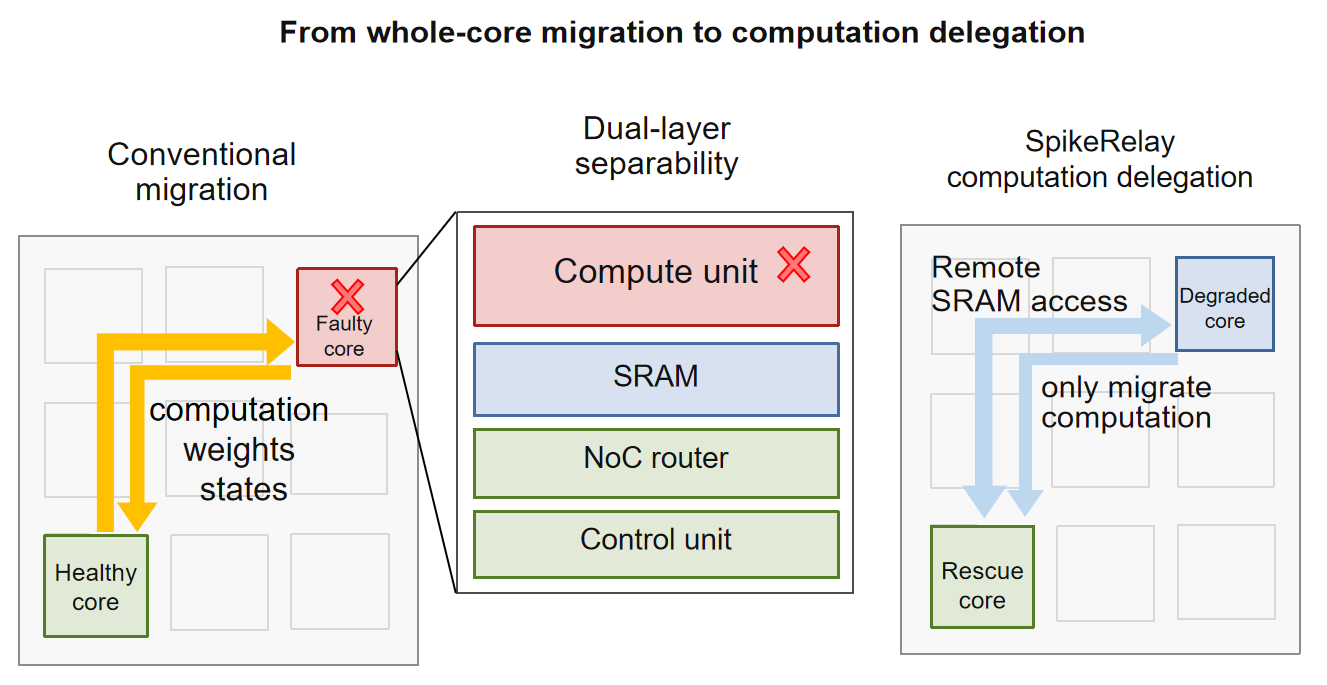}
  \caption{Conventional migration retires the entire faulty core and copies all neuron data to a healthy core, causing capacity bottlenecks at high utilization while abandoning faulty-core SRAM.}
  \label{fig:bottleneck}
\end{figure}

This design preserves the four paths required by synchronous SNN execution, so that the recovered network behaves exactly as it did before the fault. First, the rescue core can read the same weights and membrane states from the faulty core’s SRAM. Second, forwarded spike events deliver the same timestep input set to the rescue core. Third, updated membrane states are written back to their original locations, preserving temporal continuity. Fourth, output spikes are emitted with the same logical neuron identity and timestep semantics as in the baseline deployment. Under this diagnosed fault model and synchronous timestep execution, remote execution is step-equivalent to local execution. The cost of avoiding recovery-time data migration is steady-state remote memory access, which SpikeRelay controls through batched state prefetching, double buffering, and coalesced write-back.

To validate the effectiveness of SpikeRelay, we evaluate it on a Loihi-class 1,024-core simulator using three SNN workloads and dense stress scenarios. Across realistic high-utilization workloads, SpikeRelay maintains complete recovery even when migration-based baselines fail because of insufficient destination SRAM. At 92.2\% SRAM utilization and a 50\% accumulated-fault stress test, SpikeRelay achieves 100\% recovery, compared with 22.1\% for the best baseline.


In summary, the main contributions of this paper are as follows:

\begin{itemize}[nosep,leftmargin=*]

\item We identify and formalize the structural alignment between hardware modules and execution sub-tasks in neuromorphic cores, and prove that under a compute-unit fault, computation alone can be reallocated to a healthy core while storage and communication remain on the faulty core. This analysis motivates a \emph{migrate computation, not data} recovery principle.

\item We design SpikeRelay, a runtime recovery framework that implements this principle. SpikeRelay keeps the faulty core online as a remote storage node, uses $O(1)$ NoC forwarding and compact spike buffering to preserve timestep semantics, and applies batched prefetching with double buffering to hide remote-access latency. These mechanisms reduce recovery from MB-scale data migration to byte-scale metadata configuration.
  
\item We experimentally show that SpikeRelay eliminates the destination-SRAM bottleneck associated with migration-based recovery. Under high SRAM utilization, SpikeRelay achieves complete recovery in all tested fault scenarios, preserves inference accuracy, and keeps the optimized runtime overhead within the timestep budget.

\end{itemize}
\section{Background}
\label{sec:background}

\subsection{SNN Execution on Neuromorphic Systems}
\label{sec:core-arch}

Neuromorphic processors such as Loihi~1/2~\cite{davies2018loihi,davies2021advancing} and SpiNNaker~2~\cite{pedersen2019spinnaker} are organized as arrays of near-memory cores interconnected by a Network-on-Chip (NoC) that carries both inter-core spikes and access to on-chip memory~\cite{furber2014spinnaker,davies2018loihi}. As shown in Figure~\ref{fig:bottleneck}, each core integrates local SRAM, a compute unit, a control unit, and a NoC router. Although these platforms differ in scale and microarchitecture, they follow a similar execution model for SNN.

Before execution, an SNN is partitioned across cores. Each core stores the synaptic weights and dynamic states of its assigned neurons in local SRAM, while spikes are exchanged through the NoC. Execution proceeds in discrete timesteps. Under the Leaky Integrate-and-Fire (LIF) model~\cite{davies2018loihi}, the compute unit reads SRAM-resident neuron states and synaptic weights, integrates incoming spikes, applies leakage, updates membrane potentials, and emits output spikes when firing thresholds are reached.

\begin{figure}[t]
\centering
\includegraphics[width=0.7\columnwidth]{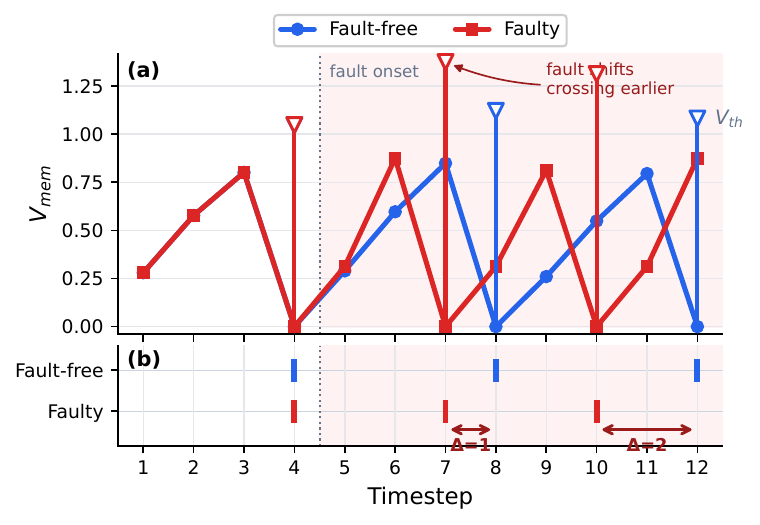}
\caption{Neuron-level impact of a stuck-at-1 fault in the compute-unit ALU. After fault onset, membrane updates diverge, threshold crossings shift, and spike displacement increases over time.}
\label{fig:fault-spike-deviation}
\end{figure}
This model has two properties central to fault recovery. First, storage, computation, and communication are physically distinct: neuron context resides in SRAM, neuron updates are performed by the compute unit, and spikes are delivered by the NoC. 
Second, one core must hold the synaptic weights and dynamic states of all of them in its SRAM. Recovering a faulty core therefore amounts to reconstructing this entire dataflow: the computation must regain access to the weights and states of every affected neuron, together with the spikes destined for them.

\subsection{The Impact of compute unit fualts}
\label{sec:fault-background}
Long deployment lifetimes can induce device aging and eventually produce permanent faults in the arithmetic logic of the compute unit~\cite{borkar2006designing,srinivasan2004impact,song2020aging,guerrero2023understanding,guerrero2025effective,li2008trace}. Unlike transient errors, permanent faults persist across timesteps and power cycles, repeatedly corrupting neuron updates once activated.

This work focuses on diagnosed and isolated permanent compute-unit faults, in which local neuron computation becomes unreliable while the affected core's SRAM contents and NoC connectivity remain intact. For faults that also corrupt storage or communication, we fall back to whole-core retirement and rely on conventional redundancy, repair, or remapping mechanisms.

A compute-unit fault perturbs membrane-potential evolution even under identical inputs. Figure~\ref{fig:fault-spike-deviation} shows a stuck-at fault that introduces a persistent update bias, advances threshold crossings, changes reset timing, and produces a spike train different from fault-free execution. Because membrane potential carries temporal state, the error accumulates across timesteps rather than remaining a one-time perturbation. The resulting shifted or missing spikes can further perturb downstream neurons, gradually altering network dynamics.

\subsection{Recovery Granularity and Architectural Opportunity}
\label{sec:recovery-opportunity}

A common response to permanent compute-unit faults is to retire the affected core and remap its neurons to healthy cores~\cite{putra2021respawn,putra2023rescuesnn,liu2021hyca}. This core-level strategy treats the faulty core as an indivisible resource, retiring its computation, weights, states, and communication all at once. Recovery is then constrained by storage rather than computation, because migrating a neuron relocates not just its update but its entire SRAM-resident context. When the original mapping already consumes most SRAM, healthy cores may still have spare compute cycles yet lack the capacity to host migrated weights and states, so recovery fails even though usable storage remains on the faulty core and healthy computation remains elsewhere.

This failure reflects a mismatch between fault scope and recovery granularity. A compute-unit fault disables reliable local updates, yet it neither invalidates the SRAM-resident neuron context nor breaks NoC reachability. The opportunity is therefore to keep neuron data in place on the faulty core and redirect only the missing computation to healthy datapaths. In the next section, we prove that, for the isolated compute-unit faults we target, redirecting only the computation can make the recovered network reproduce the original execution exactly.

\section{System Model}
\label{sec:system-model}

This section describes the system abstraction and fault model behind SpikeRelay. 
The key observation is that computation, storage, and communication are implemented by separate hardware modules, allowing recovery to relocate only the failed computation.

\subsection{System Model and Task Separation}

We consider a many-core neuromorphic system $\mathcal{P}$ consisting of $M$ cores interconnected through a Network-on-Chip (NoC). The neuron set $\mathcal{N}$ is partitioned across cores, and each core stores the weights and states of its assigned neurons locally.

Each core is modeled as
$C_i=\langle \mathcal{A}_i,\mathit{CU}_i,\mathcal{M}_i,\mathcal{R}_i\rangle$,
where $\mathcal{A}_i$ performs neuron updates, $\mathcal{M}_i$ stores neuron states and synaptic data, and $\mathcal{R}_i$ handles spike communication.

\begin{figure}[t]
  \centering
  \includegraphics[width=\columnwidth]{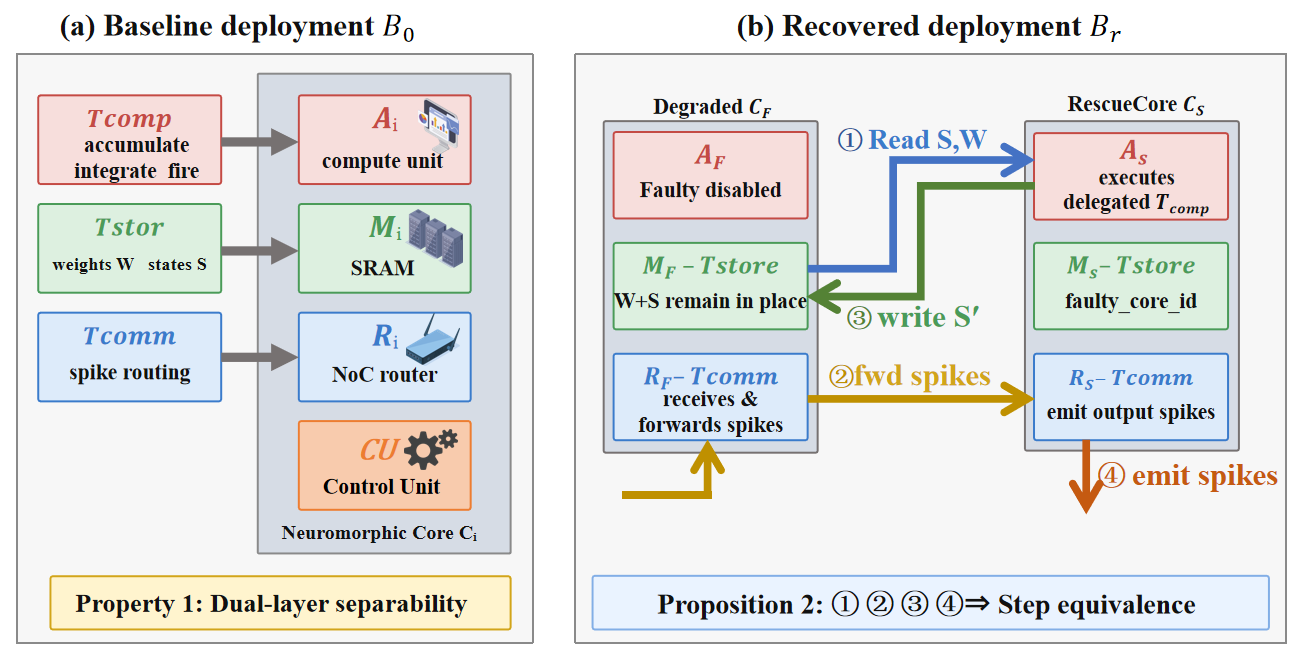}
  \caption{System model and post-fault deployment rebinding. A compute-unit fault disables only computation. SpikeRelay rebinds $\mathcal{T}_{comp}$ to a healthy rescue core while keeping storage and communication on the faulty core.}
  \label{fig:system-model}
\end{figure}

As shown in Figure~\ref{fig:system-model}, SNN execution can be decomposed into three tasks: computation ($\mathcal{T}_{comp}$), storage ($\mathcal{T}_{stor}$), and communication ($\mathcal{T}_{comm}$). These tasks are mapped to the compute unit, SRAM, and NoC router, respectively.

We refer to this alignment between hardware modules and execution tasks as \emph{dual-layer separability}. It provides the foundation for selective recovery: when only the compute unit fails, the neuron data stored in SRAM and the communication path through the router remain available.

\subsection{Fault Model and Recovery Semantics}

This work focuses on diagnosed, in-field permanent faults in the compute unit caused by device aging. We assume that the faulty core's SRAM and NoC router remain functional, while the compute unit can no longer correctly execute neuron updates.

Let $C_F$ denote a faulty core and $C_S$ denote a healthy rescue core. Instead of migrating all neuron data from $C_F$, SpikeRelay rebinds only $\mathcal{T}_{comp}$ to $C_S$. The recovered execution preserves correctness as long as the rescue core can access the original neuron states and weights, receive the same input spikes, write updated states back to the original locations, and emit spikes with the same logical identity.

Therefore, selective migration transforms recovery from a data relocation problem into a computation reassignment problem.

\section{SpikeRelay System Design}
\label{sec:design}

Building on the observation that compute, storage, and communication are separable in neuromorphic cores, we design SpikeRelay, a runtime recovery mechanism for permanent compute-unit faults.

Unlike conventional recovery approaches that retire a faulty core and migrate its neuron data, SpikeRelay keeps the faulty core online as a storage and communication node. The missing computation is delegated to a healthy rescue core, which remotely accesses the original neuron states and weights, performs neuron updates, and writes the updated states back. Therefore, SpikeRelay migrates computation rather than data.

\begin{figure}[t]
  \centering
  \includegraphics[width=\columnwidth]{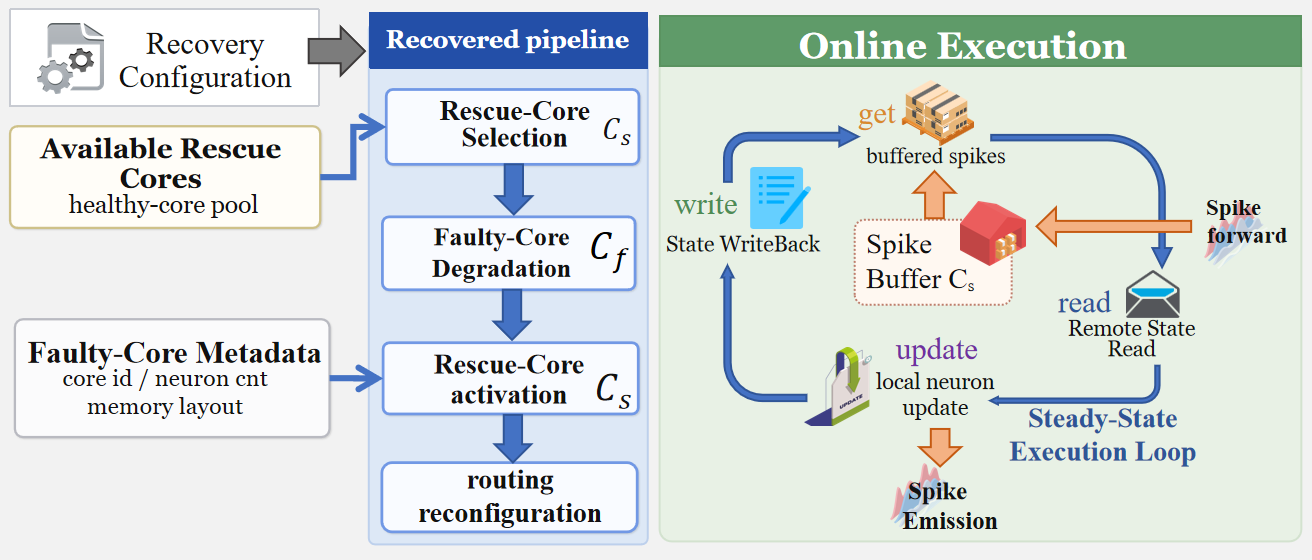}
\caption{Overview of SpikeRelay. Left and middle: recovery configuration constructs the recovered pipeline via rescue-core selection, faulty-core degradation, rescue-core activation, and routing reconfiguration. Right: during online execution, the rescue core accumulates forwarded spikes, reads the state from the faulty core, performs local updates, writes the updated state back, and emits output spikes. Only computation is migrated; weights and states remain on the faulty core.}
  \label{fig:spikerelay-overview}
\end{figure}

Figure~\ref{fig:spikerelay-overview} illustrates the recovery workflow. Recovery consists of two phases: a one-time configuration phase and steady-state remote execution. During configuration, SpikeRelay selects a rescue core, disables faulty computation, and installs forwarding rules without moving neuron data. During execution, incoming spikes are redirected to the rescue core, which performs the neuron update while the faulty core continues providing storage and communication services.
\subsection{Overview}
After recovery, affected neurons become remote neurons: their architectural state remains stored in the faulty core, while their computation executes on the rescue core. This separation eliminates the storage bottleneck caused by data migration.

\subsection{Recovery Configuration}

When a compute-unit fault is detected, SpikeRelay reconstructs the deployment through control-plane reconfiguration rather than data movement. The recovery process performs three operations:

(1) selecting a healthy rescue core according to available resources;
(2) switching the faulty core into a storage-and-forwarding mode;
and (3) configuring the rescue core and NoC forwarding path.

These operations only modify control metadata and routing states. The weights, neuron states, and synaptic data remain in their original locations throughout recovery. Therefore, recovery latency depends on the amount of configuration information rather than the size of the affected model data.

\subsection{Remote Execution}

During normal execution after recovery, the rescue core performs neuron updates for affected neurons using remote data access. It receives forwarded spikes from the faulty core, reads the required neuron states and weights through the NoC, executes the same LIF update as the original core, and writes updated states back to the faulty core.

To reduce the overhead introduced by remote accesses, SpikeRelay employs three lightweight optimizations: batch prefetching, double buffering, and write coalescing. These techniques overlap communication with computation and reduce the number of NoC transactions.

The resulting execution preserves the original neuron state transition while avoiding recovery-time migration of model data. SpikeRelay therefore converts recovery from a data relocation problem into a computation reassignment problem.

\section{Experimental Evaluation}
\label{sec:evaluation}

\subsection{Experimental Setup}
\label{sec:setup}

\noindent\textbf{Simulator and Workloads.}
We evaluate SpikeRelay on a neuromorphic simulator modeled after Intel Loihi~\cite{davies2018loihi, davies2021advancing}, consisting of a $32\times32$ core array (1,024 cores with 768\,KB SRAM per core). The simulator models neuron execution, SRAM constraints, and cycle-level NoC communication.

We evaluate three real SNN workloads and dense synthetic stress cases, covering SRAM utilizations from 73.3\% to ${\sim}$99.1\%: (a)~Spiking VGG-11 on CIFAR-100 (${\sim}$58k neurons, 73.3\%)\cite{sengupta2019going}; (b)~CIFAR10-Conv5 on CIFAR-10 (${\sim}$77k neurons, 85.8\%)\cite{wu2019direct}; (c)~SpikeBERT on SST-2 (${\sim}$84k neurons, 92.2\%)\cite{lv2025spikebert}; and dense fully-connected deployments at 95\% and ${\sim}$99.1\% utilization for stress and scalability studies. This range reflects realistic deployment because SNN mapping tools~\cite{balaji2019mapping,jin2025mapping} pack neurons tightly and production models already reach $>$90\% utilization, leaving exactly the scarce headroom in which migration-based recovery is most likely to fail. Each experiment runs 512 timesteps of inference. Permanent compute-unit faults are injected uniformly at random among deployed cores at rates from 0\% to 50\%, and each configuration is repeated with 5 random seeds.

\noindent\textbf{Baselines.}
We compare SpikeRelay against four representative baselines spanning the main recovery philosophies under a unified neuromorphic simulator abstraction: (1)~RescueSNN~\cite{putra2023rescuesnn}, a fault-aware remapping baseline that relocates affected neurons after faults; (2)~HyCA~\cite{liu2021hyca}, adapted in our simulator as a reserved recompute-resource scheme with RCC-like spare cores at the chip periphery (6.25\% of total cores); (3)~LEA~\cite{yousuf2025layer}, implemented in our simulator as a replica-based redundancy baseline with pre-provisioned neuron replicas, relying on surviving replicas after faults; and (4)~No Recovery, which directly decommissions faulty cores.

\noindent\textbf{Metrics.}
We evaluate recovery success rate, recovery-time data migration, and runtime overhead after recovery. Recovery success rate measures the fraction of affected neurons that remain executable, while runtime overhead captures the additional execution cost introduced by remote execution.

\subsection{Recovery Effectiveness}
\label{sec:recovery}

\subsubsection{Recovery Success Rate.}
\label{sec:recovery-rate}

\begin{figure}[t]
  \centering
  \includegraphics[width=0.8\linewidth]{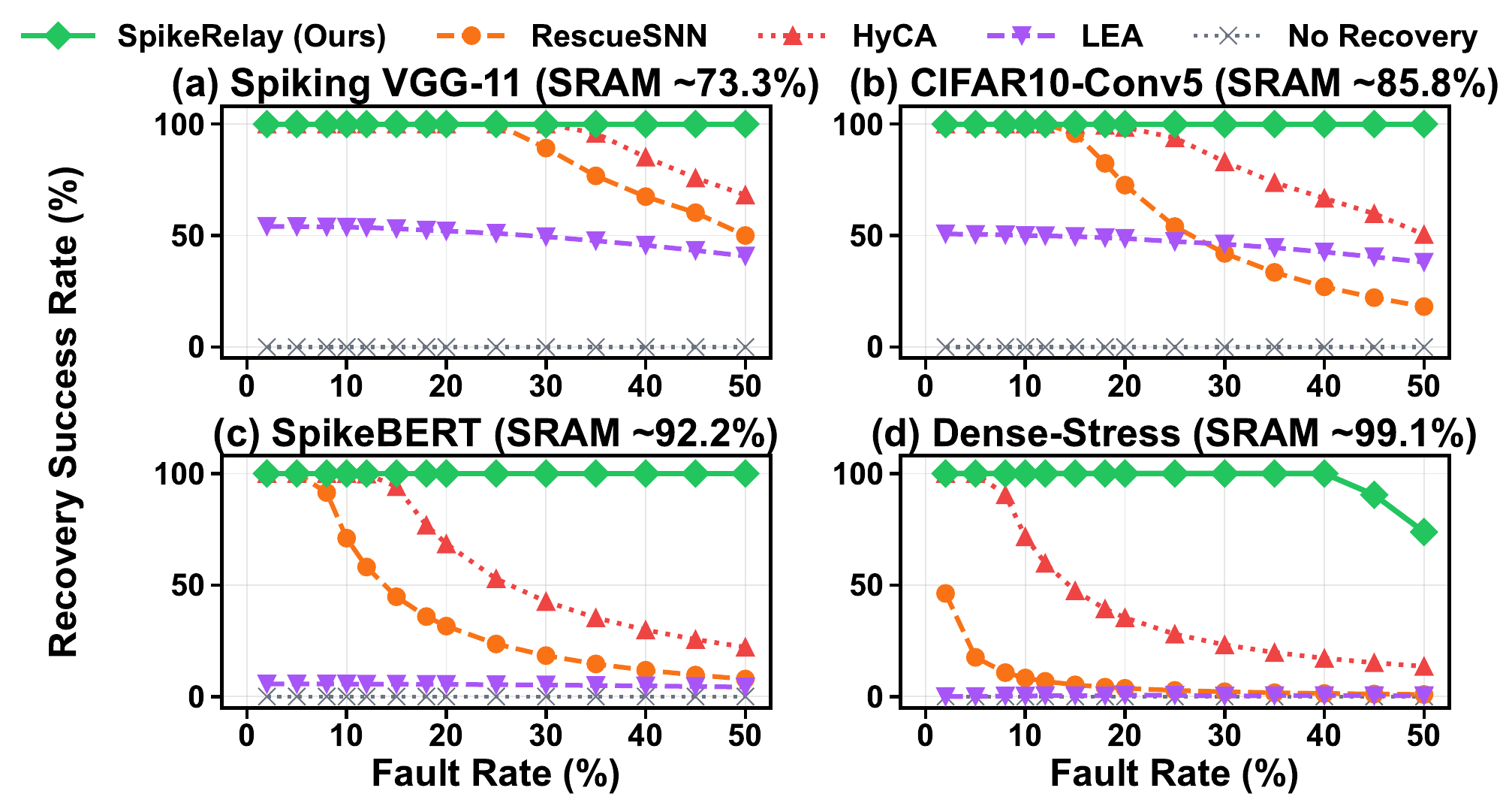}
  \caption{Recovery success rate vs.\ fault rate on a $32{\times}32$ chip across three real workloads and an extreme Dense-Stress case. SpikeRelay maintains 100\% recovery on real workloads, whereas migration-based strategies degrade as utilization increases.}
  \label{fig:recovery}
\end{figure}

Figure~\ref{fig:recovery} reports recovery success rates for the three real workloads and the extreme Dense-Stress case as the fault rate increases from 0\% to 50\%. SpikeRelay achieves 100\% recovery on real workloads (73\%--92\% utilization) across all fault rates because neuron data is never relocated: faulty cores retain all parameters and are served remotely by rescue cores. By contrast, RescueSNN and HyCA must migrate affected neuron data to healthy targets after faults and therefore degrade as utilization rises. On Dense-Stress (${\sim}$99.1\%), both data-migration baselines collapse near a 5\% fault rate due to capacity limits, whereas SpikeRelay still maintains 100\% recovery up to 40\%. HyCA exhausts its fixed RCC pool early, and LEA's best-effort replicas offer little benefit under such high utilization.

SpikeRelay's own limit is the $O(N)$ spike buffer (4\,B/neuron, about 500$\times$ smaller per neuron than full migration). On Dense-Stress, the extreme 99.1\% utilization leaves only ${\sim}$7.4\,KB free per core, while each faulty core's 2,497 neurons require about ${\sim}$10\,KB of buffer space. Recovery therefore declines to ${\sim}$90\% at 45\% and ${\sim}$74\% at 50\% fault rate. Importantly, this degradation is \emph{gradual} rather than the cliff-edge collapse seen in migration-based schemes, because only the most heavily loaded rescue cores saturate. The operator therefore retains a wide and predictable safety margin before any neuron becomes unrecoverable. For workloads with utilization ${\leq}$92\%, no such saturation appears across the full 0\%--50\% range. Even in the extreme ${\sim}$99\% regime beyond 40\% fault rate, whole-core retirement remains a viable fallback. Overall, these results show that migrating computation turns storage capacity from a hard recovery constraint into a much smaller buffer overhead.

\subsubsection{Data Migration and Recovery Latency.}
\label{sec:migration-latency}

\begin{figure}[t]
  \centering
  \includegraphics[width=0.8\linewidth]{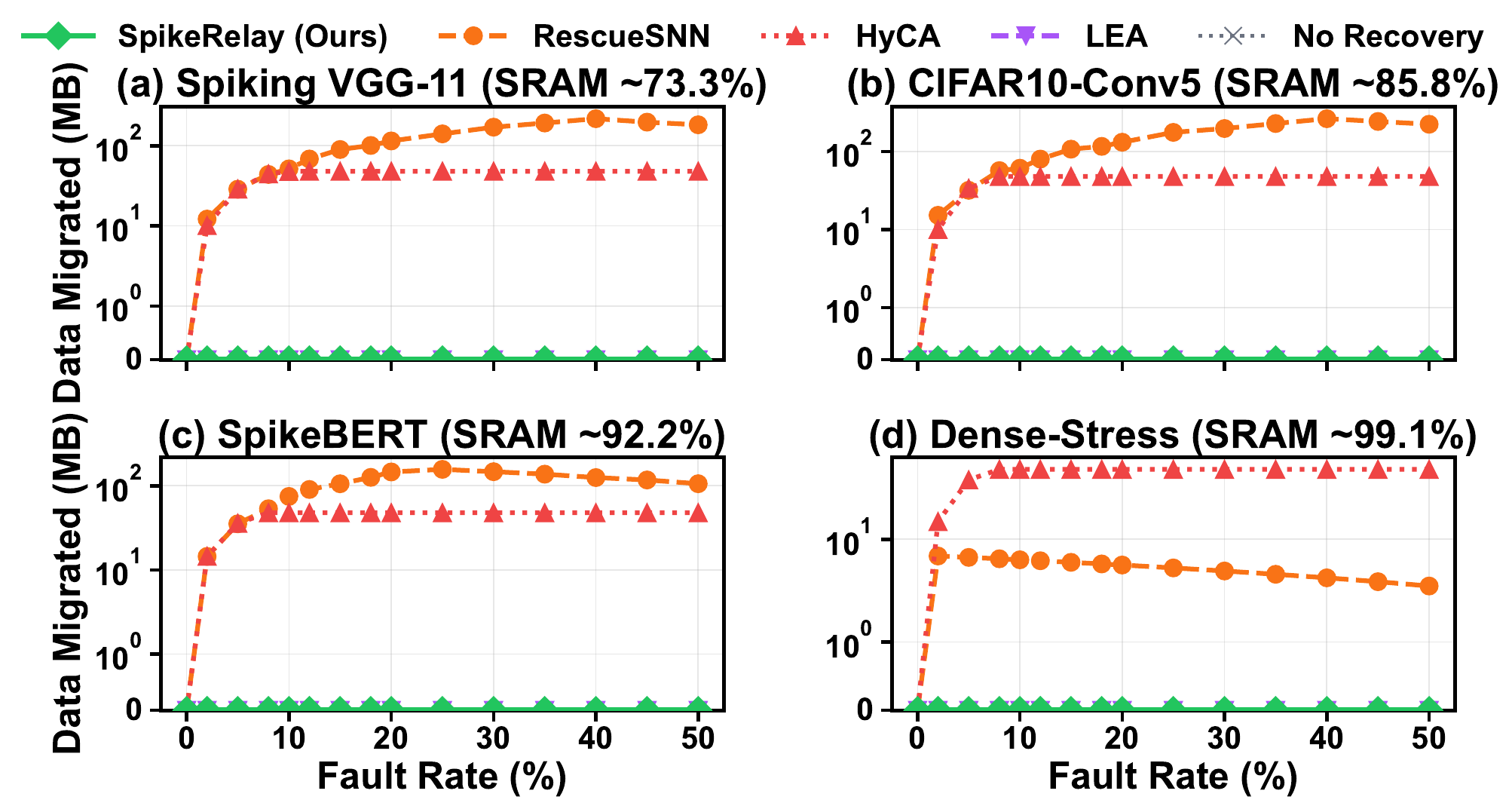}
  \caption{Data migration overhead vs.\ fault rate across the same four workloads. SpikeRelay incurs zero migration, whereas migration baselines peak and then fall as target capacity is exhausted.}
  \label{fig:migration}
\end{figure}

Figure~\ref{fig:migration} reports data migration overhead. SpikeRelay incurs no data migration at any fault rate because it keeps neuron data on faulty cores and requires only $O(1)$ metadata per recovery. Migration baselines exhibit a characteristic rise-then-fall pattern: transfer volume first grows with fault rate, then drops once target capacity is exhausted, rather than due to recovery becoming more efficient. RescueSNN exceeds 150\,MB on SpikeBERT (${\sim}$92\%) at around 25\% fault rate; on Dense-Stress (${\sim}$99\%), healthy cores fill almost immediately, so all migration baselines can barely transfer any data. LEA also incurs zero migration by design, but at the cost of near-zero recovery (\S\ref{sec:recovery-rate}). These results show that migration overhead scales with both model size and utilization, whereas SpikeRelay removes this coupling by keeping data in place.

\begin{figure}[t]
  \centering
  \includegraphics[width=0.8\linewidth]{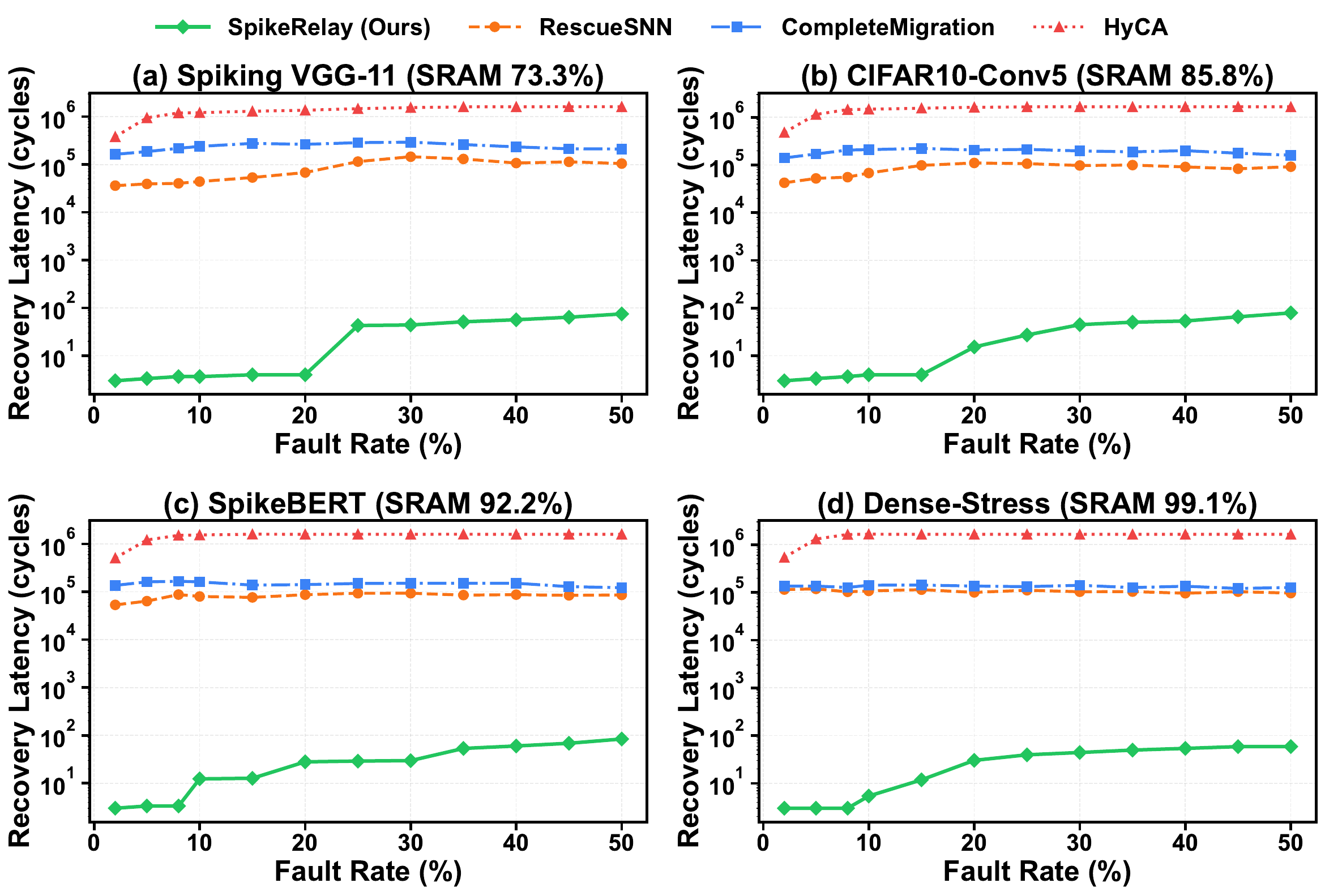}
  \caption{Cycle-accurate recovery latency vs.\ fault rate across the same four workloads. SpikeRelay completes recovery in only tens of cycles because it sends only constant-size metadata, whereas migration baselines require $10^4$--$10^6$ cycles due to bulk data movement and NoC contention.}
  \label{fig:recovery-latency}
\end{figure}

Figure~\ref{fig:recovery-latency} reports cycle-accurate recovery latency. Because SpikeRelay sends only constant-size metadata rather than neuron states, it completes recovery within 3--97 cycles on the three real workloads and 3--58 cycles on Dense-Stress across the full fault-rate range. By contrast, RescueSNN and CompleteMigration remain in the $10^4$--$10^5$-cycle regime, while the HyCA-inspired adapted baseline remains around $10^6$ cycles due to bulk-state relocation and the resulting source/destination serialization and NoC contention. Small local fluctuations in SpikeRelay's curve arise from independent random fault layouts rather than systematic sensitivity to fault rate. The gap is not merely one of volume: even on a highly parallel mesh, a faulty core's entire neuron state must be drained through a single local SRAM read port before any parallel network transfer can help, so the baselines' latency floor is set by source-port serialization rather than by hop count, and a faster NoC alone cannot close it. SpikeRelay sidesteps this serialization entirely by leaving the state in place and reading it incrementally during execution. Together with Figure~\ref{fig:migration}, these results show that migrating computation removes both the data-movement cost and the recovery-latency bottleneck of post-fault data relocation.

\subsubsection{End-to-End Inference Accuracy Validation.}
\label{sec:accuracy}

\begin{figure}[t]
  \centering
  \includegraphics[width=0.8\columnwidth]{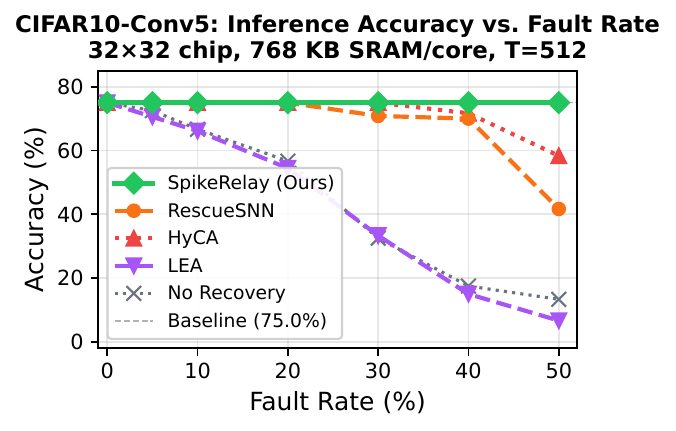}
  \caption{End-to-end inference accuracy on CIFAR10-Conv5 under 0\%--50\% fault rates. SpikeRelay maintains accuracy, while migration baselines degrade as neuron loss accumulates.}
  \label{fig:accuracy}
\end{figure}

Figure~\ref{fig:accuracy} connects recovery success rates to end-to-end task performance on CIFAR10-Conv5. SpikeRelay maintains near-constant accuracy across the full fault-rate range because 100\% recovery (\S\ref{sec:recovery-rate}) preserves the complete network structure.Additionally, LEA even underperforms No Recovery at high fault rates, as its weight pre-scaling persists after replicas are lost. In contrast, migration baselines degrade sharply beyond 20\%--25\% fault rate as neuron loss accumulates, and their accuracy closely tracks recovery rate. This confirms that incomplete recovery directly degrades accuracy, whereas SpikeRelay avoids this failure mode through remote execution, thereby empirically validating the accuracy-equivalence argument established in the system model.

\subsection{Summary}
\label{sec:eval-summary}

Across all tested configurations, in-place execution makes the five metrics reinforce rather than trade off, simultaneously driving recovery to 100\% on realistic workloads (and degrading only gradually under extreme ${\sim}$99\% utilization), migration to zero, and latency down by orders of magnitude, while accuracy tracks recovery and runtime cost stays within the real-time budget after optimization. These conclusions are consistent from 64 to 4{,}096 cores, 30\% to 95\% utilization, varying fan-out, and non-uniform fault placement.


\section{Related Work}
\label{sec:related}
Prior fault-tolerance techniques for general-purpose processors either address different fault classes or assume structures absent from neuromorphic cores. Checkpoint/rollback cannot recover permanent compute-unit failures because SNN temporal state is not fully reconstructible~\cite{sorin2002safetynet}; pipeline recombination assumes redundant stages unavailable in neuromorphic cores~\cite{powell2009architectural, gupta2010stagenet}; TMR incurs $2$--$3\times$ area overhead~\cite{lyons1962use}; fault-tolerant NoC routing handles communication faults rather than compute-unit failures~\cite{werner2016survey}; and near-memory schemes such as layer-integrated averaging target storage-cell faults rather than cases where computation fails but co-located storage survives~\cite{yousuf2025layer}.

Within neuromorphic systems, prior recovery methods fall into three groups. Reserved redundancy, including TMR, spare cores, and HyCA~\cite{liu2021hyca}, incurs upfront area and power cost. Software compensation, such as fault-aware retraining and threshold adjustment~\cite{putra2023rescuesnn, siddique2023improving}, improves tolerance but cannot restore lost compute. Runtime migration~\cite{putra2023rescuesnn, dang2022migspike, song2021dynamic} relies on whole-core retirement and copying weights and states to healthy cores, creating a destination-capacity bottleneck at high utilization and leaving usable SRAM stranded on faulty cores. SpikeRelay instead migrates only computation, keeps data in place, reduces recovery to $O(1)$ metadata configuration, and retains faulty cores as storage nodes; to our knowledge, it is the first to treat a failed compute unit's surviving storage and router as reusable resources rather than collateral loss. Table~\ref{tab:qualitative} summarizes these differences.
differences.

\begin{table}[t]
  \centering
  \caption{Qualitative comparison of fault recovery approaches.}
  \label{tab:qualitative}
  \setlength{\tabcolsep}{3pt}
  \small
  \resizebox{\columnwidth}{!}{%
  \begin{tabular}{lcccc}
    \toprule
    \textbf{Approach} & \textbf{Data Migr.} & \textbf{Latency} & \textbf{High-Util.} & \textbf{Faulty SRAM} \\
    \midrule
    HW Redund. & None & Very low & Limited & Abandoned \\
    Retrain / Adj. & None & Offline & Limited & Abandoned \\
    Data Reloc. & High & High & Limited & Abandoned \\
    \textbf{SpikeRelay} & \textbf{Zero} & \textbf{Very low} & \textbf{Yes} & \textbf{Retained} \\
    \bottomrule
  \end{tabular}
  }
\end{table}

\section{Discussion}
\label{sec:discussion}

\noindent\textbf{Fault model scope.}
SpikeRelay targets permanent compute-unit failures and composes with SRAM and NoC protection. If those modules fail beyond repair, the system retires the affected core, while SpikeRelay continues on the remaining cores. The main remaining constraint is the $4N$-byte spike buffer on~$C_S$; under extreme utilization and fault rates, multi-core partitioning may be needed, although $k{=}2$ suffices even at 99\% utilization in our experiments.

\noindent\textbf{Generality.}
Zero data migration, $O(1)$ recovery, and $O(N)$ buffer overhead stem from the design rather than device-level assumptions, though hardware validation on platforms such as Loihi~2 is still needed to quantify integration cost. Even after optimization, bottleneck-core overhead remains about 27\% at a 50\% fault rate, indicating that denser workloads, tighter real-time targets, or on-chip learning may require stronger load balancing and write-order support. These figures also assume contention-free NoC transport and are thus a lower bound that a contention-aware hardware-level evaluation should refine.

\noindent\textbf{Broader applicability.}
The key insight extends beyond neuromorphic hardware: tightly coupled compute and storage need not fail together, suggesting similar opportunities in PIM and CIM, where co-located memory likewise survives data path faults and could be served in place rather than abandoned. The current weighted-accumulator buffer assumes synchronous timesteps; asynchronous event-driven architectures~\cite{davies2021advancing} would need event-ordered queues, likely trading part of the $10{,}000\times$ buffer reduction for temporal fidelity.
\section{Conclusion}
\label{sec:conclusion}

This paper presents SpikeRelay, a fault recovery framework for many-core neuromorphic systems built on a simple observation: a compute-unit failure need not condemn an entire core. By exploiting dual-layer separability, SpikeRelay migrates computation rather than data, keeping weights and states in place and serving them remotely from healthy rescue cores. Routing redirection, standardized remote access, and compact rescue metadata replace costly bulk migration with lightweight, constant-size configuration. Experiments confirm full recovery even under high memory utilization and severe fault rates, with zero post-fault data relocation, orders-of-magnitude lower recovery latency, and runtime overhead within real-time budgets. More broadly, SpikeRelay reframes a compute-unit fault from the loss of an entire core to the loss of a single function, turning a hard capacity wall into a tunable performance trade-off.



\bibliographystyle{ACM-Reference-Format}
\bibliography{references}

\clearpage
\end{document}